\documentclass[10pt,twocolumn,letterpaper]{article}

\usepackage[pagenumbers]{cvpr} 

\makeatletter\let\captiontemp\@makecaption\makeatother
\usepackage[font=small]{subcaption}
\makeatletter\let\@makecaption\captiontemp\makeatother
\usepackage{graphicx}
\usepackage{tabularx}
\usepackage{amsmath}
\usepackage{amssymb}
\usepackage{amsthm}
\usepackage{booktabs}
\usepackage{sg-ibs-macros} 
 \usepackage{array, diagbox}
 \usepackage{pifont}
 \usepackage{stackengine}

\usepackage[accsupp]{axessibility}  

\newcolumntype{M}[1]{>{\centering\arraybackslash}m{#1}}

\newcommand{\erf}{\mathop{\textrm{erf}}}
\newcommand{\ddu}[2]{\frac{\text{d} #1}{\text{d}#2}}
\newcommand{\du}[1]{\text{d}#1}

\definecolor{darkred}{rgb}{0.7,0.1,0.1}
\definecolor{darkgreen}{rgb}{0.1,0.7,0.1}
\definecolor{dblue}{rgb}{0.2,0.2,0.8}
\definecolor{maroon}{rgb}{0.76,.13,.28}
\definecolor{burntorange}{rgb}{0.81,.33,0}
\definecolor{cyan}{rgb}{0.0,0.7,0.94}
\definecolor{salmon}{rgb}{0.99,0.51,0.46}
\definecolor{green}{rgb}{0.03,0.91,0.43}

\ifdefined\ShowNotes
  \newcommand{\colornote}[3]{{\color{#1}\bf{#2: #3}\normalfont}}
\else
  \newcommand{\colornote}[3]{}
\fi

\newcommand{\variance}[1]{\ensuremath{\textbf{Var}\left[#1\right]}}
\newcommand{\varwrt}[2][]{\ensuremath{\textbf{Var}_{#1}\!\left[#2\right]}}

\usepackage[pagebackref,breaklinks,colorlinks,allcolors=cvprblue]{hyperref}
\usepackage[capitalize]{cleveref}
\crefname{section}{Sec.}{Secs.}
\Crefname{section}{Section}{Sections}
\Crefname{table}{Table}{Tables}
\crefname{table}{Tab.}{Tabs.}

\definecolor{cvprblue}{rgb}{0.21,0.49,0.74}

\def\paperID{11534} 
\def\confName{CVPR}
\def\confYear{2025}

\usepackage{overpic}
\usepackage{enumitem} 
\usepackage{overpic} 
\usepackage{multirow}
\usepackage{colortbl}
\definecolor{grayshade}{gray}{0.8}

\definecolor{turquoise}{cmyk}{0.65,0,0.1,0.3}
\definecolor{purple}{rgb}{0.65,0,0.65}
\definecolor{dark_green}{rgb}{0, 0.5, 0}
\definecolor{orange}{rgb}{0.8, 0.6, 0.2}
\definecolor{red}{rgb}{0.8, 0.2, 0.2}
\definecolor{darkred}{rgb}{0.6, 0.1, 0.05}
\definecolor{blueish}{rgb}{0.0, 0.3, .6}
\definecolor{light_gray}{rgb}{0.7, 0.7, .7}
\definecolor{pink}{rgb}{1, 0, 1}
\definecolor{greyblue}{rgb}{0.25, 0.25, 1}

\usepackage{blindtext}

\renewcommand{\paragraph}[1]{\vspace{1em}\noindent\textbf{#1}.}
\begin{document}

\title{VI$^3$NR: Variance Informed Initialization for Implicit Neural Representation
}
\author{
  Chamin Hewa Koneputugodage$^{1}$\thanks{Equal contribution.\vspace{-0.5cm}} \quad Yizhak Ben-Shabat$^{1,2}$\footnotemark[1] \quad Sameera Ramasinghe$^{3}$\quad Stephen Gould$^{1}$\vspace{3mm}\\
  { $^{1}$The Australian National University \quad $^{2}$Roblox \quad $^{3}$Pluralis AI}\\
  {\tt \small \{chamin.hewa, stephen.gould\}@anu.edu.au sitzikbs@gmail.com sameera@pluralis.ai}\\
\tt\small { \href{https://chumbyte.github.io/vi3nr-site/}{https://chumbyte.github.io/vi3nr-site/}}\\
}


\maketitle
\begin{abstract}
Implicit Neural Representations (INRs) are a versatile and powerful tool for encoding various forms of data, including images, videos, sound, and 3D shapes. A critical factor in the success of INRs is the initialization of the network, which can significantly impact the convergence and accuracy of the learned model. Unfortunately, commonly used neural network initializations are not widely applicable for many activation functions, especially those used by INRs. In this paper, we improve upon previous initialization methods by deriving an initialization that has stable variance across layers, and applies to any activation function. We show that this generalizes many previous initialization methods, and has even better stability for well studied activations. We also show that our initialization leads to improved results with INR activation functions in multiple signal modalities. Our approach is particularly effective for Gaussian INRs, where we demonstrate that the theory of our initialization matches with task performance in multiple experiments, allowing us to achieve improvements in image, audio, and 3D surface reconstruction.
\end{abstract}


\section{Introduction}
\label{Sec:intro}
Implicit neural representations (INRs) encode diverse signals, including images, videos, 3D models, and audio, as continuous functions rather than discrete samples \cite{sitzmann2020siren, tancik2020fourier}. Given a coordinate input, a fully connected network predicts the signal value at that location. INRs excel at fitting complex, high-dimensional signals, making them ideal for 3D reconstruction \cite{park2019deepsdf, mescheder2019occupancy,atzmon2020sal,Atzmon_2020_CVPR_SALD,gropp2020igr, ben2022digs}, novel-view synthesis \cite{mildenhall2021nerf, barron2021mipnerf}, neural rendering \cite{tewari2020stateneuralrendering}, and pixel alignment \cite{saito2019pifu}.

Despite their power, INR performance depends heavily on the choice of activation function, affecting optimization, signal fitting, and learning stability. Although previous works have extensively explored the choice of activation function \cite{sitzmann2020siren,ramasinghe2022beyond,saragadam2023wire,saratchandran2024sampling}, introducing new ones requires careful weight initialization. Proper initialization is essential to preserve key properties, such as the distribution of values between layers and control of gradient magnitude. Without this careful balance, the network's effectiveness in capturing complex signals and converging efficiently may be significantly compromised.

Prior work on initialization \cite{lecun2002efficient,Glorot2010UnderstandingTD,He2015DelvingDI} has largely focused on deep learning models with sigmoid or ReLU activations for classification. These initialization techniques have become the default choice in many models, often without reconsidering their underlying assumptions. However, INRs differ significantly in focus, as they are centered on signal reconstruction rather than generalizability. In this context, the ability to capture high-frequency information is crucial, making it necessary to modify the activation functions. As a result, many of the assumptions of previous initialization approaches no longer hold, and new initialization methods are required to ensure effective learning.

We propose a novel multilayer perceptions (MLP) initialization that generalizes many previous methods \cite{lecun2002efficient,Glorot2010UnderstandingTD,He2015DelvingDI,sitzmann2020siren,kumar2017weightinitializationdeepneural} with minimal assumptions, making it applicable to any activation function. 
Similar to prior work, it requires computing activation statistics, which is difficult to do analytically. We show, however, that they can be efficiently estimated empirically, outperforming prior approximations. We evaluate the effectiveness of our proposed initialization across diverse applications, including image, audio, and 3D surface reconstruction.

The key contributions of our work are as follows: 
\begin{itemize}
    \item We propose a principled weight initialization for INRs that preserves preactivation distribution across layers and allows control of that distribution.
    \item We derive a condition for stable gradient variance across layers, guiding distribution selection for deep networks.
    \item We analyze two promising but unstable INR activation functions (Gaussian and sinc), demonstrating that our initialization significantly improves their stability and performance in multiple reconstruction tasks.
\end{itemize}

\section{Related Work}
\label{Sec:related-work}

\noindent\textbf{Neural Network Initialization. } 
Neural network initialization aims to prevent exploding or vanishing values at initialization during forward and backward passes \cite{lecun2002efficient, Glorot2010UnderstandingTD, He2015DelvingDI}. 
To ensure this, many methods initialize weights so that the variance of the preactivations (and possibly their gradients) is the same or stable across layers, with Xavier \cite{Glorot2010UnderstandingTD} and Kaiming \cite{He2015DelvingDI} initialization becoming the de facto standard.
However, Xavier relies on assumptions specific to tanh, while Kaiming is derived for ReLU. PyTorch \cite{Paszke2019PyTorch} extends these via heuristic ``gain'' adjustments, which, as discussed in \cref{sec:method:deriving-other-inits}, are not well-defined. Kumar \cite{kumar2017weightinitializationdeepneural} extends forward-pass derivations to arbitrary activations but use restrictive assumptions (unit preactivation variance) and approximations (Taylor expansions). SIREN \cite{sitzmann2020siren} derives initialization for sine activations without approximations but only for a specific input distribution. They also notice empirically that their gradients  demonstrate identical distributions, but do not formally analyze this observation.  

We build on SIREN’s formulation, extending it to arbitrary activations and distributions. Our forward-pass derivation aligns with Kumar but is more rigorous and general. We also use Monte Carlo estimation rather than Taylor approximations for computing activation statistics, yielding higher accuracy. Moreover, we generalize Xavier and Kaiming initialization by properly analyzing the backward pass, allowing us to jointly satisfy both conditions.

\noindent\textbf{Implicit Neural Representations. } 
Implicit Neural Representations (INRs) are continuous function approximators, typically based on MLPs. This continuous nature proves advantageous in various reconstruction tasks, especially when working with irregularly sampled signals like point clouds \cite{park2019deepsdf, atzmon2020sal, gropp2020igr} but also when working with regular signals like images, videos, and NeRFs \cite{sitzmann2020siren, lindell2022bacon, ramasinghe2022beyond}. However, standard MLPs often struggle to effectively capture high-frequency signals \cite{tancik2020fourier, sitzmann2020siren}. To address this, frequency encoding methods are commonly employed, where frequencies are explicitly introduced to the network. This can be achieved via Fourier feature encoding \cite{mildenhall2020nerf, tancik2020fourier} or through specific activation functions such as sinusoidal \cite{sitzmann2020siren}, variable-periodic \cite{liu2024finer}, Gaussian \cite{ramasinghe2022beyond}, wavelet \cite{saragadam2023wire}, and sinc \cite{saratchandran2024sampling}. However, these methods are sensitive to the initialization and the initial frequency bandwidth and often lack precise bandwidth control \cite{ben2022digs, lindell2022bacon}. Recent advancements address these limitations by initializing with a broad bandwidth and gradually narrowing it during optimization, either through regularization \cite{ben2022digs} or explicit frequency decomposition \cite{yifan2022geometryconsistent, lindell2022bacon}. In this work, we take a fundamental approach that demonstrates the importance of the INR initialization and propose analytical and empirical formulations that provides a performance boost for any chosen activation. 

\noindent\textbf{INR initialization. } 
In recent years, multiple works proposed initialization methods for INRs. SAL \cite{atzmon2020sal} proposed a geometric initialization that carefully selects the network weights in a way that initializes the SDF to yield a 3D sphere. SIREN \cite{sitzmann2020siren} showed that without carefully initializing the network, sine activations are not able to properly reconstruct signals. DiGS \cite{ben2022digs} proposed the multi-frequency geometric initialization (MFGI) and essentially extended the geometric initialization to sine activations and initialized to a 3D sphere while retaining high frequencies.
The initializations above are highly focused on specific cases (i.e., geometry for SAL and DiGS and sine activation for SIREN). This narrows the formulation and (other than SIREN) overlooks fundamental aspects that include the distribution of pre and post activation values for various activation functions. In this work, we show that maintaining the distribution through proper initialization results in a significant boost in performance.

\section{Approach}
\label{Sec:approach}

\begin{figure*}
    \centering
    \includegraphics[width=0.9\linewidth,trim={30 100 30 15}, clip]{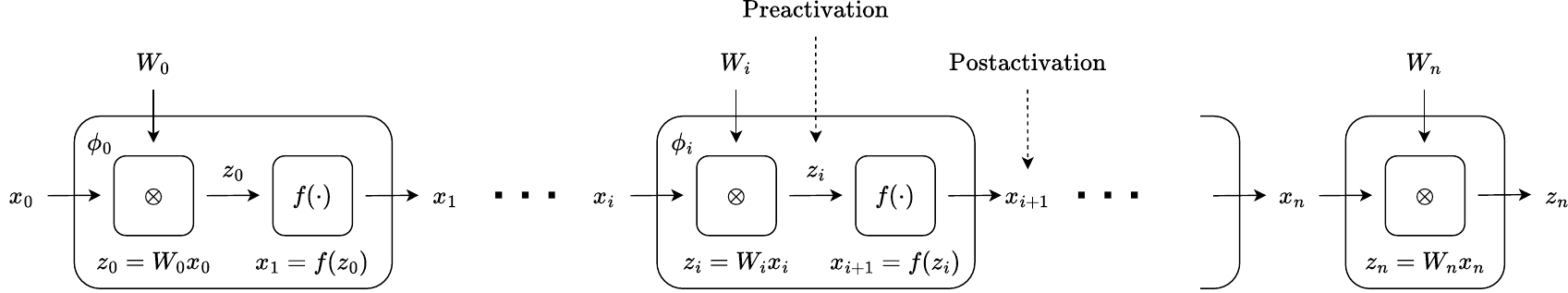}
    \caption{
    \textbf{Notation diagram.} For clarity, we illustrate a general MLP based INR architecture (layers $0$, $i$ and $n$ shown). Layers are of the form $\phi_i(x_i)=f(W_ix_i)$ where $f$ is the activation function and $W_i$ are the weights, the preactivations are $z_0,...,z_{n-1}$, the postactivations are $x_1,...,x_n$ and the input and output of the network are $x_0$ and $z_n$.    
    }
    \label{fig:network}
\end{figure*}

\subsection{Notation and Background} \label{sec:method:notation}

Following the notation of SIREN \cite{sitzmann2020siren}, we define an $n$-layer INR network $\Phi$ (also illustrated in \cref{fig:network}) as
\begin{align}
    &\Phi(x_0) = W_n\left(\phi_{n-1}\circ\phi_{n-2}\circ\ldots\circ\phi_{0}\right)(x_0) \\
    &x_i\mapsto\phi_i(x_i) = f(W_ix_i+b_i)
\end{align}
where $\phi_i:\reals^{M_i}\to\reals^{N_i}$ is the $i^\text{th}$ layer of the network so $W_i\in\reals^{N_i\times M_i}$, $b_i\in\reals^{N_i}$, $x_i\in\reals^{M_i}$ and $N_i=M_{i+1}$. Let us assume that $b_i$ is initialized to be approximately zero, and denote $z_i= W_ix_i$.

We assume that the input to the network, $x_0\in\reals^{M_0}$, is sampled from an input distribution $\D_\text{in}$ (\eg$\D_\text{in}=\mathcal{U}([0,1]^3)$ is typical for surface reconstruction). The output of the network is $z_n$, and we consider $z_0,z_1,\ldots z_{n-1}$ to be the preactivations of the network and $x_1, \ldots, x_n$ as the postactivations. We use $[\cdot]_{j}$ to index elements of matrices or vectors, so $[x_0]_j$ is the $j^\text{th}$ element of the vector $x_0$.

Finally, since we are considering vectors and matrices that have elements from the same (scalar) distribution, we will refer to the distribution of individual elements of a vector or
matrix $A$ as $\D_{A}$ with mean $\mu(A)$ and variance $\sigma^2(A)$.

\subsection{General Scheme Introduction}\label{sec:method:scheme-intro}

Our initialization is based on the observation that, under standard weight initialization assumptions, if the layer width is large enough, then the distribution of the preactivations will be Gaussian distributed with mean zero by the central limit theorem (see Prop~\ref{thm:main-prop} for formal statement). As a result, in order to ensure the preactivations have the same distribution, we only need to match the variance. Previous methods \cite{sitzmann2020siren,kumar2017weightinitializationdeepneural} implicitly assume a variance of one. However, we consider the distribution of the preactivations as a choice to specify for our initialization, as depending on the activation function the choice of preactivation distribution matters (which we investigate in \cref{Sec:results}). We also derive conditions for maintaining equal variance of gradients through the backward pass, which becomes a restriction on what preactivation distribution to actually use, though this is only relevant for very deep networks.
Our initialization requires statistics related to the activation function that can be difficult to derive analytically (depending on the activation function). We propose using Monte Carlo estimation, which we show is efficient and accurate, especially compared to previous approximations \cite{Glorot2010UnderstandingTD, kumar2017weightinitializationdeepneural}. 

\subsection{Initialization for Forward Pass} \label{sec:method:forward-pass}

We now formalize our method.
We only make the assumption that the elements of $W_i$ are generated i.i.d. from some distribution $\D_i$ with mean 0, and we set the biases, $b_i \approx 0$. Then, we prove that the preactivation distribution tends to a normal distribution for large enough layer width.
\begin{PROPOSITION}\label{thm:main-prop}
    Let the elements of $W_i\in\reals^{N_i\times M_i}$ be sampled i.i.d. from some distribution $\D_{W_i}$ with mean 0 and variance $\sigma^2(W_i)$, and the elements of $x_i$ be sampled from some distribution $\D_{x_i}$ with mean $\mu(x_i)$ and variance $\sigma^2(x_i)$. Then the elements of $z_i$, given by $[z_i]_k = \sum_{l=1}^{M_i} [W_i]_{kl}[x_i]_l$, are independent. Furthermore, as $M_i\to\infty$ the random variables $[z_i]_k$ converge in distribution to 
    \begin{align}\label{eq:preac-distr-result}
        \N\left(0, M_i\left(\mu^2(x_i)+\sigma^2(x_i)\right)\sigma^2(W_i)\right).
    \end{align}
\end{PROPOSITION}
\begin{proof}
    Note that the same (randomly drawn) vector $x_i$ is used for each $[z_i]_k$. Thus we can consider the $[z_i]_k$ to be draws from a finite weighted sum of i.i.d. random variables $[W_i]_{kl}$, so they are independent. Furthermore we can consider each product to be a random variable $p_{kl}=[W_i]_{kl}[x_i]_l$, which are random weightings of the original i.i.d. random variables $[x_i]_l$. The $p_{kl}$ are therefore independent with mean 0 but not identically distributed.    
    
    Thus, by the Central Limit Theorem with Lindeberg's condition~\cite{ash2000probability}, as $M_i\to\infty$ we have that $\frac{1}{s_{M_i}}\sum_{l=1}^{M_i} \left(p_{kl}-\expectation{p_{kl}}\right) = \frac{1}{s_{M_i}}\sum_{l=1}^{M_i}[W_i]_{kl}[x_i]_l$ converges in distribution to $\N\left(0, 1\right)$ where 
    $s_{M_i}^2:=\sum_{l=1}^{M_i} \variance{p_{kl}} = \left(\sum_{l=1}^{M_i} [x_i]_l^2\right) \sigma^2(W_i)$. For large $M_i$ this is approximately $M_i\expectation{[x_i]_l^2}\sigma^2(W_i)=M_i\left(\mu^2(x_i)+\sigma^2(x_i)\right)\sigma^2(W_i)$. Thus $[z_i]_k$ converges in distribution to $\N\left(0, M_i\left(\mu^2(x_i)+\sigma^2(x_i)\right)\sigma^2(W_i)\right)$.
\end{proof}

Note that SIREN \cite{sitzmann2020siren} proves the case for $\D_{x_i} = \sin^{-1}(-1,1)$ (their Lemma 1.5 and overarching Theorem 1.8) and Kumar \cite{kumar2017weightinitializationdeepneural} reports the same result as our (their Proposition 2) but only motivate independence and do not provide a formal proof.

Given this result, in order to make the distribution of preactivations the same at each layer, it suffices to just make the preactivation variances in \cref{eq:preac-distr-result} equal at each layer. Furthermore, note that since we control the distribution of each $W_i$, we can even set this variance ourselves. Hence, if want the variance of the preactivations to be $\sigma_p^2$, then at each layer we need to set the distribution of $W_i$ to have mean 0 and variance
\begin{align}\label{eq:derived-W-var}
    \sigma^2(W_i) = \frac{\sigma_p^2}{M_i\left(\mu^2(x_i)+\sigma^2(x_i)\right)}.
\end{align}
For the first layer, we have to use the mean and variance of the input distribution $\D_{x_0}=\D_\text{input}$. For later layers we can proceed by induction, assuming we have set the distribution of $W_{i-1}$ to be such that $[z_{i-1}]_k\sim N(0,\sigma_p^2)$, then
\begin{align}\label{eq:muxi}
    \mu(x_i) &= \mu(f(z_{i-1})) = \expectwrt[z\sim \N(0,\sigma^2_p)]{f(z)}\\
    \sigma^2(x_i) &= \sigma^2(f(z_{i-1})) = \varwrt[z\sim \N(0,\sigma^2_p)]{f(z)}.\label{eq:sigmaxi}
\end{align}
Note that we have removed the dependence on $i$, so these expectations only need to be computed once given $\sigma_p$ and $f$. This can be either done analytically, or using offline Monte Carlo estimation.

\paragraph{Uniform Distribution} Often we choose to initialize our weights using a symmetric uniform distribution, $\mathcal{U}[-c,c]$. This depends on parameter $c$ and has variance $\frac{1}{3}c^2$. Thus solving \cref{eq:derived-W-var} for $c$ we get for our intermediate weights,
\begin{align}
    c &= \sigma_p\sqrt{\frac{3}{M_i\left(\mu^2(x_i) + \sigma^2(x_i) \right)}}.
\end{align}

\subsection{Initialization for Backward Pass} \label{sec:method:backward-pass}

We may also want to ensure that the distribution in the backward pass is also stable, in particular to ensure the first two moments of the distribution of $\frac{\partial \mathcal{L}}{\partial z_i}$ are the same for all layers $i$. Consider, for any element $j$,
\begin{align}
    \frac{\partial \mathcal{L}}{\partial [z_{i-1}]_j}
        &= \sum_{k=1}^{N_i} \frac{\partial \mathcal{L}}{\partial [z_{i}]_k} [W_i]_{kj} f'([z_{i-1}]_{j}),
\end{align}
then (see derivations in the supplemental)
\begin{align}
    \mu\left(\frac{\partial \mathcal{L}}{\partial z_{i-1}}\right) 
        &= 0\\
    \sigma^2\left(\frac{\partial \mathcal{L}}{\partial z_{i-1}}\right)
        &= M_{i+1} \sigma^2\left(\frac{\partial \mathcal{L}}{\partial z_i}\right)\sigma^2(W_i)  \notag\\
        &\quad (\mu^2(f'(z_i))+\sigma^2(f'(z_i))).
\end{align}
Thus to make $\sigma^2\left(\frac{\partial \mathcal{L}}{\partial z_{i-1}}\right)=\sigma^2\left(\frac{\partial \mathcal{L}}{\partial z_i}\right)$ we must have that 
\begin{align}\label{eq:backwards-cond-initial}
    M_{i+1} \sigma^2(W_i) \left(\mu^2(f'(z_i)) + \sigma^2(f'(z_i))\right)=1.
\end{align}
Solving for $\sigma^2(W_i)$ gives
\begin{align}\label{eq:backwards-derived-W-var}
    \sigma^2(W_i) 
    = \frac{1}{M_{i+1}\left(\mu^2(f'(z_i)) + \sigma^2(f'(z_i))\right)}.
\end{align}

This likely evaluates to a different value than the expression given in \cref{eq:derived-W-var}.\footnote{Unless, for example, $f$ is the identity transformation.} The solution in Xavier initialization \cite{Glorot2010UnderstandingTD} is to average the two expressions (with $\sigma^2_p = 1$), and in Kaiming initialization~\cite{He2015DelvingDI} is to use either, as in both cases their assumptions make the terms similar. 

On the other hand, we have one more degree of freedom than both Xavier and Kaiming, namely $\sigma^2_p$. As such, we can equate \cref{eq:derived-W-var} and \cref{eq:backwards-cond-initial} to give
\begin{align}\label{eq:backwards-cond-final}
    \sigma_p^2 \frac{M_{i+1}}{M_i}  \frac{\mu^2(f'(z_i)) + \sigma^2(f'(z_i))}{\mu^2(x_i)+\sigma^2(x_i)} &= 1
\end{align}
and solve for $\sigma_p^2$. Note here that the terms $\mu(x_i)$, $\sigma^2(x_i)$, $\mu(f'(z_i))$, and $\sigma^2(f'(z_i))$ are expectations and variances over the distribution $\N(0,\sigma_p^2)$ and therefore are themselves functions of $\sigma^2_p$. We address this complication below.

\subsection{Using Our Initialization in Practice}\label{sec:method:init-in-practice}

\paragraph{Calculating \cref{eq:derived-W-var} and \cref{eq:backwards-cond-final} given $f$ and $\sigma_p^2$}
Computing the expectations in \cref{eq:derived-W-var} (defined in \cref{eq:muxi} and \cref{eq:sigmaxi}) is non-trivial. Xavier initialization simplifies this by assuming the activation function is is well approximated by a linear function, which makes the expectations trivial to compute. Kumar \cite{kumar2017weightinitializationdeepneural} uses a first-order Taylor approximation around zero and only considers $\sigma_p=1$. This approach requires the activation to be differentiable at zero and closely resemble its linear approximation at zero, which is not the case for many activation functions. As a result, both approximations can be quite poor, as shown in \cref{sec:method:comparing-activations}.
SIREN on the other hand analytically derive the expectations for their case ($f=\sin\left(\frac{\pi}{2}x\right)$, $\sigma_p^2=1$). We follow a similar procedure for Gaussian activations and derive these expectations analytically (\cref{eq:gaussian-analytical-sum}) for arbitrary $\sigma_p$ and $\sigma_a$ (a parameter in Gaussian activations). We refer to our initialization with analytically derived expectations as \textit{our init}.

However, for general activations, especially multimodal activations like sinc, such analytical derivations are non-trivial. Instead, we propose using Monte Carlo (MC) estimation: we apply the activation function $f$ to 1M samples from $\N(0,\sigma_p^2)$ and compute the mean and variance. This is very efficient in practice (taking milliseconds), and also quite accurate (see \cref{sec:method:comparing-activations} and \cref{tab:mc-ablation}). We estimate the expectations in \cref{eq:backwards-cond-final} in the same way. We refer to our initialization using MC estimates as \textit{our MC init}.

\paragraph{Calculating $\sigma_p^2$ from \cref{eq:backwards-cond-final}}
Since many terms in \cref{eq:backwards-cond-final} involve expectations over $\N(0,\sigma_p^2)$, solving for $\sigma_p$ analytically is non-trivial. Instead, we perform a fast grid search ($\sim$0.5s for 1k values) and select the $\sigma_p$ that makes the left-hand side of \cref{eq:backwards-cond-final} closest to one. This is done for the inner layers (the majority case), where $M_i=M_{i+1}$.

\paragraph{Choosing $\sigma_p^2$ in practice} The requirement that the backward variance is exactly equal between layers (which results in \cref{eq:backwards-cond-final}) is only important for very deep networks. In such cases, if the variance ratio between layers is not quite one then the variance in the backward pass eventually vanishes or explodes. However, for smaller networks, as is typical in INRs, vanishing or exploding is less likely, so this condition is less important. We note instead that the best choice for $\sigma_p$ is likely task specific, though highly influenced by the condition in \cref{eq:backwards-cond-final}. Thus after finding $\sigma_p$ by \cref{eq:backwards-cond-final}, which often already gives a significant performance boost over nominal values (e.g.,~$\sigma_p = 1$), we perform a small local line search to determine the best $\sigma_p$ for the task. Note that determining the best $\sigma_p$ only needs to be done once per task, not per instance. In practice, the small local search is minor compared to standard hyperparameter tuning (see \cref{sec:res:recon-perf}).

\subsection{Deriving Commonly Used Initializations} \label{sec:method:deriving-other-inits}

We show that our initialization is more general than commonly used initializations by deriving them as special cases. 

Xavier initialization works under assumption that the activation function $f$ is symmetric with unit derivative at 0, so that the preactivations hit a region of $f$ such that $f(x)\approx x$. Then $\mu(x_i)=0$ and $\sigma^2(x_i)=\sigma^2(z_{i-1})$. Thus in order to preserve variances, i.e., $\sigma^2(z_i)=\sigma^2(z_{i-1})$, they solve for $\sigma^2(z_i)=\sigma^2(x_i)$. By \cref{eq:derived-W-var} this happens when $\sigma^2(W_i)=\frac{\sigma^2(x_i)}{M_i\sigma^2(x_i)}=\frac{1}{M_i}$. Likewise, for the backwards pass by \cref{eq:backwards-derived-W-var} they want $\sigma^2(W_i)=\frac{1}{M_{i+1}}$ (as $\mu(f'(z_i))=1$ and $\sigma^2(f'(z_i))=0$ by their assumptions). Taking the average they arrive at $\sigma^2(W_i)=\frac{2}{M_i+M_{i+1}}$

Kaiming initialization considers ReLU activations with $\sigma_p=1$ (though their derivation is not tied to this), and notes that $\mu^2(x_i)+\sigma^2(x_i)=\mu(x_i^2)=\frac{1}{2}\sigma^2(z_{i-1})=\frac{1}{2}$ and $\mu^2(f'(z_i))+\sigma^2(f'(z_i))=\mu^2(f'(z_i))=\frac{1}{2}$. Thus by \cref{eq:derived-W-var} they arrive at $\sigma^2(W_i)=\frac{2}{M_i}$ and by \cref{eq:backwards-derived-W-var} they arrive at $\sigma^2(W_i)=\frac{2}{M_{i+1}}$. 
Kaiming initialization instead proposes to use either of these rather than their average, noting that in their case both work well. PyTorch \cite{Paszke2019PyTorch} calls the former \emph{fan-in} mode and the latter \emph{fan-out} mode. 

PyTorch generalizes both Xavier and Kaiming by introducing the ``gain'' of activation functions, $\sigma^2(W_i)=\text{gain}^2(f)\frac{2}{M_i+M_{i+1}}$ for Xavier and $\sigma^2(W_i)=\text{gain}^2(f)\frac{1}{M_i}$ for Kaiming. While never explictly defined, it is motivated as a scaling term on the weight's variance to compensate for the activation function, which implies
\begin{align}\label{eq:orig-gain}
    \text{gain}^2(f) &\approx \frac{\sigma^2(z_i)}{\sigma^2(f(z_i))}.
\end{align}
However this is not well defined as it could vary between layers, and thus raises the question of what should be used for $\sigma^2(z_i)$. Often gain is computed with $\sigma^2(z_i)=1$, but it is not made clear to users that the gain value is only valid if they ensure that their first layer preactivations have that variance. The other approach is to treat the gain as a hyperparameter and brute force search for a value that keeps the variance through the network stable. Note that our initialization gives a well defined meaning to this value:
\begin{align}\label{eq:our-gain}
    \text{gain}^2(f, \sigma_p) &= \frac{\sigma_p^2}{\expectwrt[z]{f(z)}^2+\varwrt[z]{f(z)}}
\end{align}
where $z\sim \N(0,\sigma^2_p)$.

\subsection{Applying to Specific Activations} \label{sec:method:comparing-activations}

\begin{table}
    \small
    \centering
    \begin{tabular}{l | c c c}
        \textbf{Method} & \textbf{tanh} & \textbf{sigmoid} &
        \textbf{ReLU}\\ 
        \toprule 
        Xavier \cite{Glorot2010UnderstandingTD} &
            $\frac{1}{M_i}$ & - & - \\
        Kaiming \cite{He2015DelvingDI} &
            - & - & $\frac{2}{M_i}$ \\
        Kumar \cite{kumar2017weightinitializationdeepneural} &
            $\frac{1}{M_i}$ & $\frac{12.96}{M_i}$ & $\frac{2}{M_i}$ \\ 
        PyTorch \cite{Paszke2019PyTorch} &
            $\frac{2.78}{M_i}$ & $\frac{1}{M_i}$ & $\frac{2}{M_i}$ \\ 
        Our MC init &
            $\frac{2.54}{M_i}$ & $\frac{3.41}{M_i}$ & $\frac{2}{M_i}$\\
        \hline
        \rowcolor{grayshade}
        $\sigma_p$ from \cref{eq:backwards-cond-final} &
            0.1 & 6.8 & $1^*$\\
        Our MC init &
            $\frac{1.02}{M_i}$ & $\frac{104.28}{M_i}$ & $\frac{2}{M_i}$\\
    \end{tabular}
    \caption{\textbf{Different initialization's $\sigma^2_{W_i}$.} Our method yields different values for tanh and sigmoid. Interestingly, our theoretical value is close to PyTorch's experimentally chosen value for tanh. \textbf{Top}: uses $\sigma_p^2=1$, \textbf{Bottom}: chooses $\sigma_p^2$ based on condition \cref{eq:backwards-cond-final} (found using a grid search). $^*$ all $\sigma_p$ satisfy \cref{eq:backwards-cond-final}.}
    \label{tab:init-comp}
\end{table}

\begin{table}
    \small
    \setlength{\tabcolsep}{2.7pt}
    \centering
    \begin{tabular}{l | c c c c c c}
        & \multicolumn{2}{c}{\textbf{tanh}} & \multicolumn{2}{c}{\textbf{sigmoid}} & \multicolumn{2}{c}{\textbf{ReLU}}\\
        \textbf{Method} & $E_f$ & $E_b$ & $E_f$ & $E_b$ & $E_f$ & $E_b$ \\ 
        \toprule 
        Xavier \cite{Glorot2010UnderstandingTD} &
            99.0 & 98.5 & - & -  & - & - \\
        Kaiming \cite{He2015DelvingDI} &
            - & - & - & - & 9.7 & 25.9 \\
        Kumar \cite{kumar2017weightinitializationdeepneural} &
            99.0 & 98.5 & 62.3 & 100.0 & 9.7 & 25.9 \\ 
        PyTorch \cite{Paszke2019PyTorch} &
            7.8 & 100.0 & 59.9 & 100.0 & 9.7 & 25.9 \\
        Our MC init &
            2.1 & 100.0 & 3.0 & 100.0 & 9.7 & 25.9 \\
        \hline
        \rowcolor{grayshade}
        $\sigma_p$ from \cref{eq:backwards-cond-final} &
            \multicolumn{2}{c}{0.1} & \multicolumn{2}{c}{6.8} & \multicolumn{2}{c}{1}\\
        Our MC init &
            8.8 & 21.1 & 3.6 & 39.5 & 9.7 & 25.9\\
    \end{tabular}
    \caption{\textbf{Controlled initialization method experiment.} Reporting errors in the forward ($E_f$) and backward ($E_b$) pass for different initialization methods. Our initialization provides lower forward errors when $\sigma_p^2=1$ (\textbf{Top}), and lower forward and backward errors when we choose $\sigma_p^2$ using the condition in \cref{eq:backwards-cond-final}  (\textbf{Bottom}).}
    \label{tab:init-res}
\end{table}

\begin{table}
    \small
    \setlength{\tabcolsep}{2.7pt}
    \centering
    \begin{tabular}{l | c c c | c c c c}
         &  \multicolumn{3}{c}{$\sigma_p=1$} &  \multicolumn{4}{c}{$\sigma_p$ from \cref{eq:backwards-cond-final}}\\ 
        \textbf{Activation} & $\sigma^2(W_i)$ & $E_f$ & $E_b$ & $\sigma_p$ & $\sigma^2(W_i)$ & $E_f$ & $E_b$\\ 
        \toprule 
        Sine
            & $\frac{2}{M_i}$ & 2.0 & 100.0 
            & 0.004 & $\frac{0.001}{M_i}$ & 10.1 & 20.4\\
        Gaussian
            & $\frac{28.07}{M_i}$ & 6.7 & 100.0 
            & 0.078 & $\frac{0.015}{M_i}$ & 0.9 & 35.6\\
        Sinc
            & $\frac{1.31}{M_i}$ & 3.2 & 100.0
            & 2.225 & $\frac{10.700}{M_i}$ & 0.3 & 21.4\\
        Wavelet
            & $\frac{2.68}{M_i}$ & 1.7 & 100.0
            & 0.871 &  $\frac{1.805}{M_i}$ & 0.8 & 22.5\\
    \end{tabular}
    \caption{\textbf{INR activation experiment.} We follow same setup as \cref{tab:init-res} for our MC initialization. For all activation functions, when $\sigma_p^2=1$ we achieve low forward errors, and when we choose $\sigma_p^2$ using the condition in \cref{eq:backwards-cond-final} we achieve low forward and backward errors. For Wavelet we use the real part of a Gabor wavelet.}
    \label{tab:inr-init-res}
\end{table}

\paragraph{Common MLP activations}
In \cref{tab:init-comp}, we compare the values of $\sigma^2(W_i)$ produced by different initialization methods. For simplicity, we assume $M_i = M_{i+1}$, so the forward and backward conditions are the same for Xavier and Kaiming. We use the Monte Carlo (MC) version of our initialization. 
In the top part of the table, we fix $\sigma_p = 1$, following the common assumption in prior work. Existing initialization methods use approximations for tanh and sigmoid, but not for ReLU. Consequently, their ReLU initializations match ours. PyTorch, however, uses an empirically derived value for tanh, which happens to closely align with our initialization.
In the bottom part, we set $\sigma_p^2$ by grid-searching to satisfy the backwards condition in \cref{eq:backwards-cond-final}.

\cref{tab:init-res} presents a controlled evaluation of these values, roughly following the setup used in PyTorch's tests \cite{Paszke2019PyTorch}. We construct a 100-layer MLP with 1000 units per layer. The network input is a preactivation with variance $\sigma_p^2$, and we measure the error $E_f$ between the preactivation variance at the last layer and $\sigma_p^2$ using SMAPE. For the backward pass, we propagate gradients with unit variance and compute the error $E_b$ between the resulting input gradient variance and one (also using SMAPE).
Results show that for $\sigma_p = 1$, our method yields equal or lower forward-pass error than others, while all methods perform similarly in the backward pass, showing high error for tanh and sigmoid, and low error for ReLU. Notably, Xavier and Kumar exhibit high forward-pass error due to poor first-order Taylor approximations. PyTorch’s empirically derived tanh value performs well in the forward pass. When using $\sigma_p$ selected via our backwards condition, our method achieves low error in both the forward and backward passes.

\paragraph{INR activations}
We evaluate INR-specific activations in \cref{tab:inr-init-res}. Using $\sigma_p = 1$ results in low forward error, but high backward error, while $\sigma_p$ based on \cref{eq:backwards-cond-final} (via grid search) yields low forward and backward error. We found these activations were highly sensitive to initialization, thus we performed the grid search with finer resolution to ensure accuracy. For each activation, we use standard forms: $\sin(ax)$ with $a=30$ (sine), $\exp\left(-x^2 / 2\sigma_a^2\right)$ with $\sigma_a=0.05$ (Gaussian), $\frac{\sin(ax)}{ax}$ with $a=1$ (sinc), and $\cos(ax)\exp(-(ax)^2)$ with $a=1$ (wavelet). Note that our initialization for sine is the same as in SIREN \cite{sitzmann2020siren}. We further discuss SIREN's initialization in the supplemental.

\paragraph{Default initialization}
When an activation function does not have an initialization derived for it, either general random noise is used, or Xavier/Kaiming is used. However as the latter are derived for other activation functions, this is a misuse. However, we sometimes find that is the activation function has a parameter that can be tuned, then the activation function itself can be tuned to make Xavier/Kaiming work. We demonstrate this in \cref{sec:res:xavier-kaiming-comp}.

\section{Experiments}
\label{Sec:results}
We use three INR activation functions: sine, Gaussian and sinc. As sine activations is well established in the literature by SIREN \cite{sitzmann2020siren}, we use their chosen scaling, $f(x)=\sin(30x)$, along with their initialization (which is equivalent to ours for $\sigma_p=1$). For Gaussian and sinc activations we do not fix their activation parameters ($\sigma_a$ and $a$). 
 For Gaussians, we compute the expectations in \cref{eq:backwards-derived-W-var} both analytically (\textit{our init}) and via Monte Carlo (\textit{our MC init}). For sinc the analytic derivation is difficult due to its multimodality and only estimate expectations with MC estimation (\textit{our MC init}). 
 As a baseline we use random normal noise with fixed variance across all layers to initialize the weights (\textit{random normal init}). For all experiments we use 8 layers each with 128 units.

\paragraph{Hyperparameter search}
We perform hyperparameter tuning for both $\sigma_p$ and the activation function parameters ($\sigma_a$ for Gaussians, $a$ for sinc) for each task. Note that we do not tune for sine activations since $\omega=30$ and $\sigma_p=1$ has been shown to work well for multiple tasks in the literature~\cite{sitzmann2020siren}. Note that for Gaussians and sinc it is standard to grid search over their activation function parameter for each task \cite{ramasinghe2022beyond,saratchandran2024sampling}, which can be costly (\textbf{$\sim$5 hours}) due to the lack of a clear prior for the optimal value. 

Our initialization introduces $\sigma_p$, which naively would require a grid search for each activation function parameter. However following \cref{sec:method:init-in-practice}, for each activation function parameter we can set $\sigma_p$ directly using its theoretical value \cref{eq:backwards-cond-final}. In practice, this value is only accurate for deeper networks (\cref{fig:std_grid_search_num_layers}), but for shallower networks a small local line search, overall taking \textbf{10--20mins}, improves results. Furthermore, in many cases (e.g.,~3D SDFs in Fig.~4) the theoretical value already gives a significant performance boost over nominal values (e.g.,~$\sigma_p = 1$).
Finally, not only is the overhead introduced by our grid search over $\sigma_p$ small compared to the required grid search over $\sigma_a$, it is minor compared to standard hyperparameter tuning (e.g., learning rate selection) required in all neural approaches. Since this is an initialization hyperparameter, its effects are very clear at the start of optimization, and we find we can quickly disregard many choices of $\sigma_p$ if the loss decreases very slowly in the first few iterations optimizations.

For our random normal init baselines, we grid search both on the activation function parameters and on the variance of random normal noise used to initialise the weights.

\paragraph{Computing activation function statistics}
For Gaussians we analytically derive (see supplemental for derivation)
\begin{align}
    \mu^2(x_i) + \sigma^2(x_i) &= \frac{\sigma_r}{\sqrt{\sigma_r^2+2}} \label{eq:gaussian-analytical-sum}
\end{align}
where $\sigma_r = \frac{\sigma_a}{\sigma_p}$. Notably, it depends only on the ratio $\sigma_r$ between the two standard deviations.

To check the quality of our MC estimates, we compare these analytical expectations with the MC estimates for Gaussians with varying sample sizes. We show the error in these estimates and the time taken to compute the estimates in \cref{tab:mc-ablation}. We also show the error in the forward and backward pass as per \cref{tab:inr-init-res}. Note that the estimation quality is good for $\geq 10K$ samples, while still being quick to compute (12ms for 1M samples).


\begin{table}
    \small
    \centering
    \setlength{\tabcolsep}{4pt}
    \begin{tabular}{l | c c c c c}
        & \textbf{GT} & \textbf{1M} & \textbf{100K} & \textbf{10K} & \textbf{1K}\\
        \toprule 
        $E_W$
            & - & 7.7e-4 & 2.3e-3 & 8.1e-3 & 2.2e-2\\
        Time (ms)
            & - & 12.0 & 1.5 & 0.4 & 0.2\\
        $E_f$
            & 2.1 & 3.4 & 3.0 & 2.8 & 3.4\\
        $E_b$
            & 18.0 & 18.3 & 20.1 & 19.7 & 35.1 \\
    \end{tabular}
    \caption{\textbf{Monte Carlo ablation.} Ablation on the MC estimation quality and time for four different sample sizes with Gaussians. $E_W$: error in computing \cref{eq:derived-W-var} (GT from \cref{eq:gaussian-analytical-sum}), $E_f,E_b$: error in forward and backward pass following \cref{tab:inr-init-res}.}
    \label{tab:mc-ablation}
\end{table}


\subsection{Reconstruction performance}\label{sec:res:recon-perf}
We evaluate our initialization on three reconstruction tasks: images, 3D signed distance functions (SDFs) and audio.

\begin{table}
    \small
    \centering
    \begin{tabular}{l | c c}
        \textbf{Method} & \textbf{Mean} & \textbf{Std} \\ 
        \toprule 
            SIREN + random normal init & 10.71 & 3.11 \\
            SIREN + SIREN init & \textbf{55.77} & 2.21 \\
            \hline
            Sinc + random normal init & 32.11 & 2.73 \\
            Sinc + our empirical init & \textbf{71.44} & 5.25\\ 
            \hline
            Gaussian + random normal init  & 30.42 & 4.41\\ 
            Gaussian + Our MC init   & 75.85 & 6.26\\ 
            Gaussian + Our init             & \textbf{79.73} & 11.85 \\ 
            
    \end{tabular}
    \caption{\textbf{Image reconstruction.} PSNR results on the KODIM dataset \cite{kodaklossless}. The results show that initializing the INRs using our initialization results in significant performance gains for all activation functions, even when our MC init is used.}
    \label{tab:image-results}
\end{table}

\begin{table}
    \small
    \centering
    \begin{tabular}{l | c c}
        \textbf{Method} & \textbf{Mean} & \textbf{Std} \\ 
        \toprule 
        SIREN + random normal init & 0.0139 & 0.0071\\
        SIREN + SIREN init  & \textbf{0.9791} & 0.0112\\
        \hline
        Sinc + random normal init & 0.7252 & 0.1037 \\
        Sinc + Our MC init & \textbf{0.9055} & 0.0466 \\
        \hline
        Gaussian + random normal init  & 0.8628 & 0.0718\\
        Gaussian + Our MC init & 0.9693 & 0.0149\\
        Gaussian + Our init & \textbf{0.9697} & 0.0148\\
    \end{tabular}
    \caption{\textbf{3D SDF reconstruction.} IoU results on the shapes used in BACON \cite{lindell2022bacon}. The results show that for all activation functions, properly initializing the INRs results in significant improvement over general initialization. Note that our MC and analytical initializations achieves similar results.}
    \label{tab:sdf-results}
\end{table}

\begin{table}[tb]
    \small
    \centering
    \setlength{\tabcolsep}{2.7pt}
    \begin{tabular}{ l  c  c  c}
    \toprule
         \textbf{Architecture} &  \textbf{Bach} & \textbf{Counting} & \textbf{Two 
         Speakers }\\
         
         \toprule 
         SIREN + random normal init & 35.5 & 19.4 &30.79\\
         SIREN + SIREN init  & 0.005 & 0.37 & 0.018\\
        \hline
         Gaussian + random normal init  & 6.57 & 5.10 & 1.35 \\
         Gaussian + Our MC init & 1.17 & \textbf{1.36} & 0.25 \\
         Gaussian + Our init            & \textbf{0.48} & 1.49 & \textbf{0.24}\\
         \bottomrule
    \end{tabular}
    \caption{\textbf{Audio reconstruction}. MSE results ($\times 10^{-3}$) on audio segments from SIREN \cite{sitzmann2020siren}. The results show that the proposed initialization methods can significantly outperform random initialization where the MC and analytical performance are comparable with a slight advantage for the analytical formulation. }
    \label{tab:results:audio_recon}
\end{table}

\paragraph{Image Reconstruction} 
We evaluate image reconstruction on the KODIM dataset \cite{kodaklossless}, achieving significantly better performance than random normal baselines for both sinc and Gaussian activations, and outperforming SIREN (a competitive baseline in the literature). For Gaussians, our MC init shows only a slight performance drop.

\paragraph{3D SDF Reconstruction}
We evaluate on the BACON dataset \cite{lindell2022bacon}, which includes four shapes. Our method outperforms random normal baselines for both sinc and Gaussian activations but remains behind SIREN.

\paragraph{Audio Reconstruction}
We reconstruct three audio signals from SIREN \cite{sitzmann2020siren} and report mean squared error in \tabref{tab:results:audio_recon}. Details of the slightly different setup are in the supplemental. Our initialization outperforms random init, with a slight edge for the analytical formulation. 

\begin{figure*}
    \centering
    \begin{subfigure}{0.32\textwidth}
        \centering
        \includegraphics[width=\linewidth]{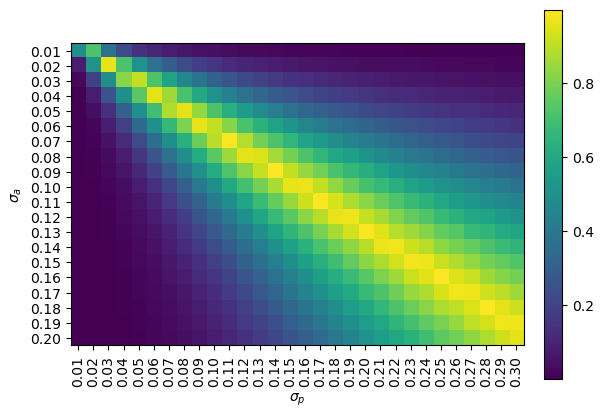}
        \caption{\textbf{Backwards condition visualization.} One minus the error in the backwards condition (\cref{eq:backwards-cond-final}) as a function of $\sigma_a$ and $\sigma_p$. This theoretically derived condition suggests a strong linear trend.}
        \label{fig:std_grid_search_backwards_cond}
    \end{subfigure}
    \hfill
    \begin{subfigure}{0.32\textwidth}
        \centering
        \includegraphics[width=\linewidth]{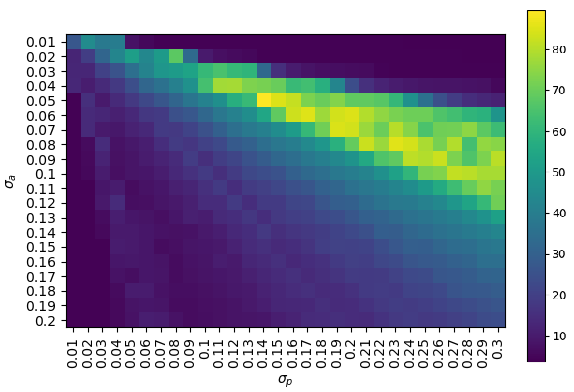}
        \caption{\textbf{Parameter grid search for images.} PSNR results as a function of $\sigma_a$ and $\sigma_p$. As the theory suggests, there is a linear trend, with some deviation from the theoretical slope in (a).}
        \label{fig:std_grid_search}
    \end{subfigure}
    \hfill
    \begin{subfigure}{0.32\textwidth}
        \centering
        \includegraphics[width=\linewidth]{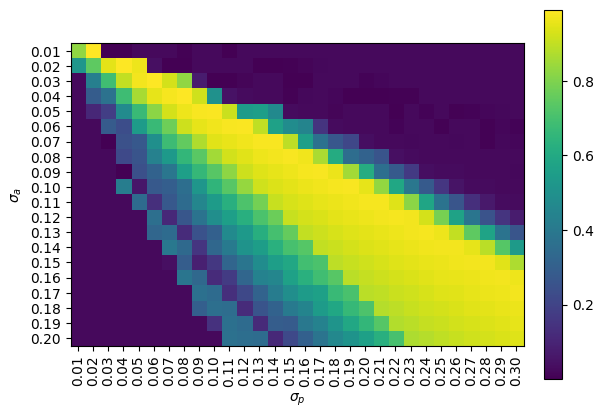}
        \caption{\textbf{Parameter grid search for 3D SDFs.} IoU results as a function of $\sigma_a$ and $\sigma_p$. As the theory suggests, there is a linear trend, with a similar slope to the theoretical slope in (a).}
        \label{fig:std_grid_search_3d}
    \end{subfigure}
    \caption{Heatmaps of the condition in \cref{eq:backwards-cond-final} and task performance when using our initialization with Gaussian activations. In each heatmap 
    the Gaussian activation function variance ($\sigma_a^2$) and preactivation variance ($\sigma_p^2$) are varied, revealing linear trends.}
    \label{fig:combined_grid_search}
\end{figure*}

\begin{figure}
    \centering
    \includegraphics[width=\columnwidth]{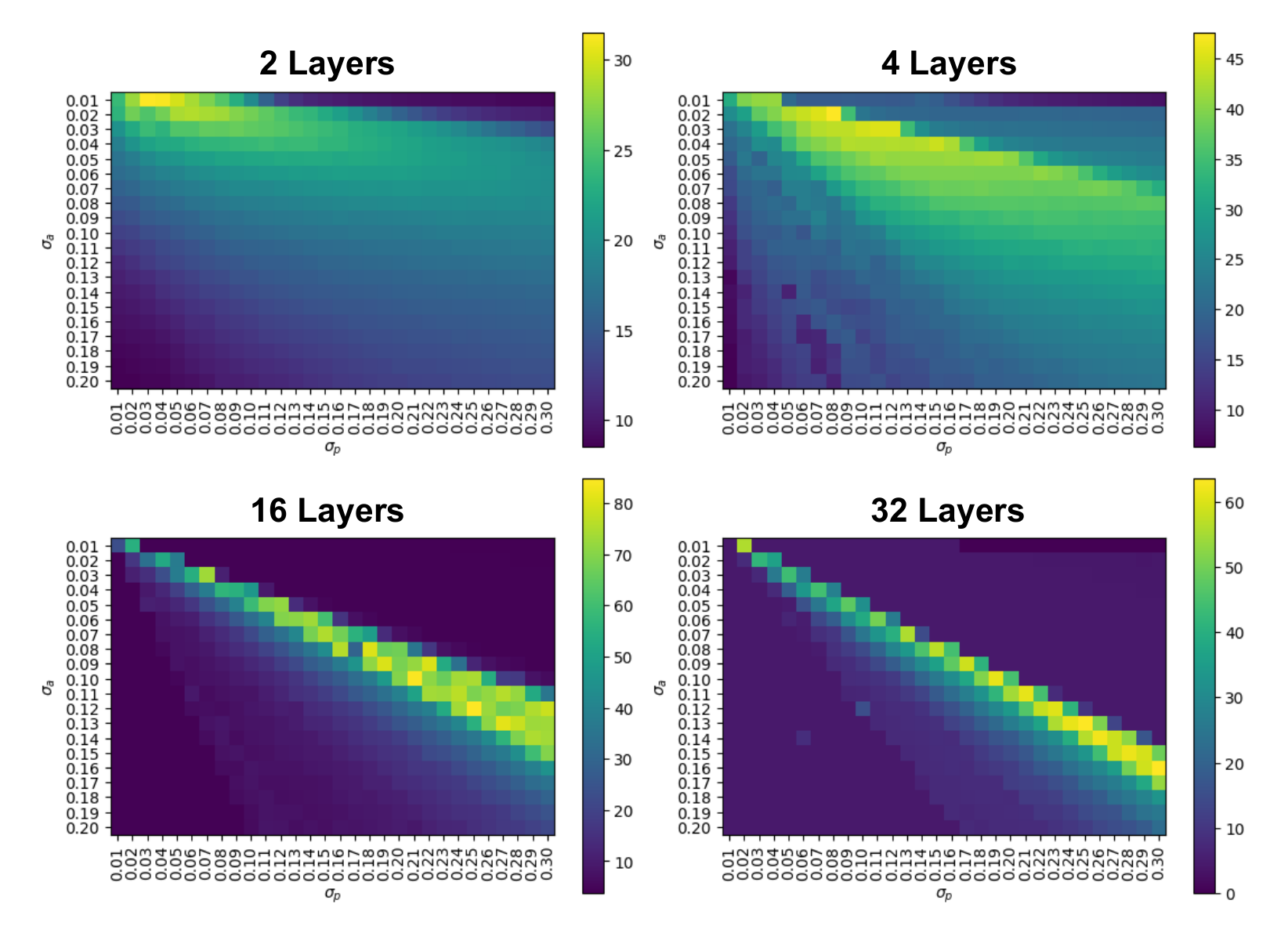}
    \caption{\textbf{Parameter grid search for images with differing number of layers.}  We plot PSNR as a function of the Gaussian activation standard deviation ($\sigma_a$) and the weight distribution standard deviation ($\sigma_p$). 
    As the number of layers increases, the slope converges towards the theoretical slope, and the linear trend sharpens. This matches our hypothesis that the backward condition \cref{eq:backwards-cond-final} is more important as the number of layers increase. \vspace{-2mm}}
    \label{fig:std_grid_search_num_layers}
\end{figure}

\subsection{Investigating Preactivation Variance}
We investigate the derived backwards condition (\cref{eq:backwards-cond-final}) with Gaussian activations by grid-searching over $\sigma_a$ and $\sigma_p$ and plotting one minus the error in the condition (using SMAPE) in \cref{fig:std_grid_search_backwards_cond}. When computing \cref{eq:backwards-cond-final}, we assume $M_i=M_{i+1}$ as this is true for most layers (all hidden layers). As expected, the optimal $\sigma_p$ depends on activation parameters. Surprisingly, we observe a strong linear trend of $\sigma_r=\frac{\sigma_a}{\sigma_p} \approx 0.66$, likely due to the expectations in \cref{eq:gaussian-analytical-sum} depending only on $\sigma_r$.

With the same grid search, we run image reconstruction and SDF reconstruction over a single instance and plot the task performance in \cref{fig:std_grid_search} and \cref{fig:std_grid_search_3d}. As shown, there is a similar linear trend as \cref{fig:std_grid_search_backwards_cond}, however for images the trend is $\sigma_r\approx 0.33$ and for SDFs the trend is $\sigma_r\approx 0.66$ but a lot wider and slightly skewed to $\sigma_r\approx 0.5$. As these are identical networks (8 layers and 256 units), this suggests that the best $\sigma_p^2$ is indeed also task specific. We investigated whether it is due to the dimension of inputs and outputs and were not able to find any trend there, suggesting it is more likely due to the task's data distribution. We ran similar grid searches over other images and other 3D SDFs to confirm that a similar trend occurs.

To test our hypothesis that the backward condition is more important as the number of layers increases, we run the same experiment for images with 2, 4, 16 and 32 layers in \cref{fig:std_grid_search_num_layers}. Here we can see that as the number of layers increase, the trend of the best performing methods gets sharper and approaches the theoretical trend in \cref{fig:std_grid_search_backwards_cond}. On the other hand, in the case of small number of layers, the trend is different, suggesting that there could be some other factors in play. For an intermediate number of layers there is a trade-off between the two trends.


\begin{table}
    \footnotesize
    \centering
    \setlength{\tabcolsep}{3pt}
    \begin{tabular}{l l l l | c c}
        \textbf{Act.} &\textbf{Init.}  & \textbf{D.F.} & \textbf{Hyperparams.} & \textbf{Mean} & \textbf{Std} \\ 
        \toprule 
            ReLU & Xavier 
                & Tanh
                &
                & 18.85 & 2.93\\
            ReLU & Kaiming 
                & ReLU
                &
                & 18.36 & 2.87\\
            ReLU & Ours 
                & Any
                & $\sigma_p=1$
                & 18.36 & 2.88\\
            \hline
            Gaussian & Xavier
                & Tanh
                & $\sigma_a=0.05$, $(\sigma_p=0.33)$
                & 14.52 & 2.70\\
            Gaussian & Kaiming
                & ReLU
                & $\sigma_a=0.05$, $(\sigma_p=0.41)$
                & 9.97 & 1.34\\
            Gaussian & Ours
                & Any
                & $\sigma_a=0.05$, $\sigma_p^\dagger=0.08$
                & 25.08 & 2.96\\
            Gaussian & Ours
                & Any
                & $\sigma_a=0.05$, $\sigma_p^*=0.15$
                & 79.73 & 11.85\\
            \hline
            Gaussian & Xavier
                & Tanh
                & $\sigma_a^*=0.09$, $(\sigma_p=0.40)$
                & 62.60 & 5.32\\
            Gaussian & Kaiming
                & ReLU
                & $\sigma_a^*=0.20$, $(\sigma_p=0.65)$
                & 79.70 & 7.68\\
            Gaussian & Ours
                & Any
                & $\sigma_a^*=0.05$, $\sigma_p^*=0.15$
                & 79.73 & 11.85\\
    \end{tabular}
    \caption{\textbf{Comparison to Xavier and Kaiming init.} PSNR results on the KODIM dataset~\cite{kodaklossless} Note Kaiming and Xavier init are equivalent to our Gaussian init for some $\sigma_p$ (put in parentheses). \textit{D.F.}: Derived for, $^\dagger$: parameter chosen using \cref{eq:backwards-cond-final}, $^*$:  parameter chosen by grid search.}
    \label{tab:kaiming-xavier-comp}
\end{table}


\subsection{Comparing to Xavier and Kaiming}\label{sec:res:xavier-kaiming-comp}
We provide comparisons to Xavier and Kaiming init on image reconstruction in \cref{tab:kaiming-xavier-comp}. We first compare with ReLU activations (which Kaiming init was derived for), which results in inferior performance as expected \cite{ramasinghe2022beyond,sitzmann2020siren}. Note that our ReLU init is exactly the same as Kaiming init (see \cref{tab:init-comp}). We then compare with Gaussian activations. This is a misuse of Xavier and Kaiming init as they were derived for tanh and ReLU respectively, but using them is common when no activation-specific init is available. Interestingly, since $\sigma_p$ is a free parameter in our init, Kaiming and Xavier are equivalent to our Gaussian init for some $\sigma_p$ (see supplemental for more details). We show this in \cref{tab:kaiming-xavier-comp} Middle ($\sigma_a=0.05$ chosen from the literature \cite{ramasinghe2022beyond}), using \cref{eq:backwards-cond-final} gives $\sigma_p=0.08$ which is close to the optimal (see \cref{fig:std_grid_search}), doing a local grid search finds the optimal $\sigma_p$ ($\sigma_p=0.15$), and Xavier and Kaiming init correspond to our init for $\sigma_p=0.33$ and $\sigma_p=0.41$ respectively. In \cref{tab:kaiming-xavier-comp} Bottom, we show that we can grid search over $\sigma_a$ to find a activation function that works well for for Xavier and Kaiming init. Note that the best-performing configurations align with the trend in \cref{fig:std_grid_search} ($\frac{\sigma_a}{\sigma_p}\approx 0.33$).


\section{Conclusion}
\label{Sec:conclusions}
In this work, we introduced a novel initialization method for INRs that effectively addresses the limitations of conventional initializations across various activation functions. Our approach generalizes existing methods by maintaining stable variance across layers, regardless of activation type, and thus supports a wider range of INR applications, including image, audio, and 3D reconstruction. We also establish a connection between the activation parameters and weight distribution parameters. Through theoretical analysis and extensive experiments, we demonstrate that our initialization enhances the stability and performance of INRs, particularly with challenging activation functions like Gaussian and sinc. Our work provides a robust foundation for future INR research, and broadens the practical applications of neural representations in encoding high-dimensional data.

\medskip
\noindent

{
    \small
    \bibliographystyle{ieeenat_fullname}
    \bibliography{references}
}


\clearpage
\setcounter{page}{1}
\maketitlesupplementary

\section{Derivations}

\subsection{Derivation for the backward pass}
Given
\begin{align}
    \frac{\partial \mathcal{L}}{\partial [z_{i-1}]_j}
        &= \sum_{k=1}^{N_i} \frac{\partial \mathcal{L}}{\partial [z_{i}]_k} [W_i]_{kj} f'([z_{i-1}]_{j}),
\end{align}
we have that
\begin{align}
    \mu\left(\frac{\partial \mathcal{L}}{\partial z_{i-1}}\right) 
        &= \sum_{k=1}^{N_n} \expectation{[W_i]_{kj}} \expectation{\frac{\partial \mathcal{L}}{\partial [z_{i}]_k} f'([z_{i-1}]_{j})}\\
        &= 0
\end{align}
and
\begin{align}
    \sigma^2\left(\frac{\partial \mathcal{L}}{\partial z_{i-1}}\right)
        &= \sum_{k=1}^{N_i} \variance{\frac{\partial \mathcal{L}}{\partial [z_{i}]_k}} \variance{[W_i]_{kj}} \notag\\
        &\quad\left(\expectation{f'([z_{i-1}]_{j})}^2 + \variance{f'([z_{i-1}]_{j})}\right)\\
        &= N_i \sigma^2\left(\frac{\partial \mathcal{L}}{\partial z_i}\right)\sigma^2(W_i)  \notag\\
        &\quad(\mu^2(f'(z_i))+\sigma^2(f'(z_i))).
\end{align}
Finally note that $N_i=M_{i+1}$.

\subsection{Initialization for the first layer}
For our INR experiments, we first normalize the input coordinates $x_0$ to be within $[-1,1]^D$ (for some input dimension $D$). We then model our element input distribution as a uniform distribution over $[-1,1]$, so $\D_\textbf{in}=\mathcal{U}([-1,1])$. Then we have that
\begin{align}
    \expectation{x_0} &= 0\\
    \textbf{Var}(x_0) &= \frac{1}{12}2^2 = \frac{1}{3}
\end{align}
so $\mu^2(x_0)+\sigma^2(x_0) = \frac{1}{3}$.
Thus by \cref{eq:derived-W-var} we initialize our first layer weights with variance
\begin{align}
    \sigma^2(W_0) 
        &= \frac{\sigma_p^2}{M_0\left(\mu^2(x_0)+\sigma^2(x_0)\right)}\\
        &= \frac{3\sigma_p^2}{M_0}
\end{align}
which is equivalent to by $\mathcal{U}([-c,c])$ where 
\begin{align}
    c 
        &= \sigma_p\sqrt{\frac{3}{M_0\left(\mu^2(x_0) + \sigma^2(x_0) \right)}}\\
        &= \sigma_p\sqrt{\frac{9}{M_0}}.
\end{align}

\subsection{Analytical Expectations for Gaussians}

Let us assume that the preactivations at layer $i-1$, $z_{i-1}$, have variance $\sigma_p^2$. Then to ensure that $z_i$ has the same variance, we set the variance of $W_i$ according to \cref{eq:derived-W-var}, which requires us to compute the mean and variance of $x_i$. We do the analytical derivation of this now.

Given $X\sim\N(0,\sigma_p^2)$ and $Y=\exp\left(\frac{-X^2}{2\sigma_a^2}\right)$ then
\begin{align}
    F_X(x) = \frac{1}{2}+\frac{1}{2}\erf\left(\frac{x}{\sigma_p\sqrt{2}}\right)
\end{align}
and $0\leq Y\leq 1$ and
\begin{align}
    F_Y(y)
        &= \pr{\exp\left(\frac{-X^2}{2\sigma_a^2}\right) \leq y}\\
        &= \pr{\frac{-X^2}{2\sigma_a^2} \leq \log(y)}\\
        &= \pr{X^2 \geq -2\sigma_a^2\log(y)}\\
        &= \pr{\abs{X} \geq \sigma_a\sqrt{-2\log(y)}}\\
        &= 2\pr{X \leq -\sigma_a\sqrt{-2\log(y)}} \tag{as $\N(0,\sigma_p^2)$ is symmetric around $0$}\\
        &= 2\left(\frac{1}{2}+\frac{1}{2}\erf\left(\frac{-\sigma_a\sqrt{-2\log(y)}}{\sigma_p\sqrt{2}}\right)\right)\\
        &= 1 + \erf\left(-\frac{\sigma_a}{\sigma_p}\sqrt{-\log(y)}\right)\\
        &= 1 + \erf\left(-\sigma_r\sqrt{-\log(y)}\right)
\end{align}
where $\sigma_r = \frac{\sigma_a}{\sigma_p}$.

Thus
\begin{align}
    f_Y(y)
        &= \ddu{}{y}F_y(y)\\
        &= \ddu{}{y}\left(1 + \erf\left(-\sigma_r\sqrt{-\log(y)}\right)\right)\\
        &= \frac{2}{\sqrt{\pi}} \exp(\sigma_r^2\log(y))\ddu{}{y}\left(-\sigma_r\sqrt{-\log(y)}\right)\\
        &= \frac{2}{\sqrt{\pi}} y^{\sigma_r^2}\frac{\sigma_r}{2y\sqrt{-\log(y)}}\\
        &= \frac{\sigma_r}{\sqrt{-\pi\log(y)}} y^{\sigma_r^2-1}.\\
\end{align}

This has mean
\begin{align}
    \expectation{y}
        &= \int_0^1 yf_Y(y)\du{y}\\
        &= \int_0^1 \frac{\sigma_r}{\sqrt{\pi}\sqrt{-\log(y)}} y^{\sigma_r^2} \ \du{y}\\
        &= \frac{\sigma_r}{\sqrt{\pi}}\int_0^1 \left(-\log(y)\right)^{-0.5} y^{\sigma_r^2} \ \du{y}\\
        &= \frac{\sigma_r}{\sqrt{\pi}} \frac{\Gamma(-0.5+1)}{\left(\sigma_r^2+1\right)^{-0.5+1}} \\
        &= \frac{\sigma_r}{\sqrt{\pi}} \frac{\sqrt{\pi}}{\sqrt{\sigma_r^2+1}}\\
        &= \frac{\sigma_r}{\sqrt{\sigma_r^2+1}}
\end{align}
and variance

\begin{align}
    \textbf{Var}(y)
        &= \int_0^1 y^2f_Y(y)\du{y} - \expectation{y}^2\\
        &= \int_0^1 \frac{\sigma_r}{\sqrt{\pi}\sqrt{-\log(y)}} y^{\sigma_r^2+1} \ \du{y} - \expectation{y}^2\\
        &= \frac{\sigma_r}{\sqrt{\sigma_r^2+2}} - \frac{\sigma_r^2}{\sigma_r^2+1}
\end{align}

so
\begin{align}
    \mu^2(x_i) + \sigma^2(x_i)
        &= \frac{\sigma_r}{\sqrt{\sigma_r^2+2}}.
\end{align}

Thus by \cref{eq:derived-W-var} we initialize our $i^\textbf{th}$ layer weights with variance
\begin{align}
    \sigma^2(W_i) 
        &= \frac{\sigma_p^2}{M_i\left(\mu^2(x_i)+\sigma^2(x_i)\right)}\\
        &= \frac{\sigma_p^2\sqrt{\sigma_r^2+2}}{M_i\sigma_r}
\end{align}
which is equivalent to by $\mathcal{U}([-c,c])$ where 
\begin{align}
    c 
        &= \sigma_p\sqrt{\frac{3}{M_i\left(\mu^2(x_i) + \sigma^2(x_i) \right)}}\\
        &= \sigma_p\sqrt{\frac{3\sqrt{\sigma_r^2+2}}{M_i\sigma_r}}.
\end{align}
\section{SIREN's initialization}
Note that SIREN \cite{sitzmann2020siren} also uses sine activations and specifically uses $\sigma_p=1$. In their derivation they use $\sin\left(\frac{\pi}{2}x\right)$ in order to only consider the monotonic region of sine. Our method also gives $\sigma^2(W_i)=\frac{2}{M_i}$ with that activation function. In the code, SIREN actually use $\sin(30x)$ which our method also gives $\sigma^2(W_i)=\frac{2}{M_i}$. However, for $\sin(x)$ our method gives $\sigma^2(W_i)=\frac{2.31}{M_i}$.

\section{Comparison to Xavier and Kaiming init.}
Our initialization for Gaussians is
\begin{align}
    \sigma^2(W_i) 
        &= \frac{\sigma_p^2}{M_i\left(\mu^2(x_i)+\sigma^2(x_i)\right)}\\
        &= \frac{\sigma_p^2\sqrt{\sigma_r^2+2}}{M_i\sigma_r}
\end{align}
where $\sigma_r=\frac{\sigma_a}{\sigma_p}$, while Xavier initialization is of the form (for middle layers)
\begin{align}
    \sigma^2(W_i) 
        &= \frac{1}{M_i}
\end{align}
and Kaiming initialization is of the form
\begin{align}
    \sigma^2(W_i) 
        &= \frac{2}{M_i}.
\end{align}
Thus for a fixed $\sigma_a$ (the Gaussian activation function parameter), Xavier and Kaiming are equivalent to our Gaussian initialization for some $\sigma_p$. For example, $\sigma_a=0.05$, $\sigma_p=0.33$ with our Gaussian init gives $\sigma^2(W_i)=\frac{1.02}{M_i}$ matching Xavier initialization, and $\sigma_a=0.05$, $\sigma_p=0.41$ gives $\sigma^2(W_i)=\frac{1.96}{M_i}$ matching Kaiming. 

For fixed $\sigma_a$ (so a fixed Gaussian activation function), it is unlikely that the $\sigma_p$ that Xavier or Kaiming correspond to are optimal (\cref{tab:kaiming-xavier-comp} Middle). However, if we grid search on $\sigma_a$ (\ie on the activation function) then it is possible that there will be a $\sigma_a$ such that Xavier and Kaiming will correspond to the optimal $\sigma_p$ for that $\sigma_a$ (\cref{tab:kaiming-xavier-comp} Bottom). In fact, the observed trend in \cref{fig:std_grid_search} makes it quite likely.

\section{Image Comparison.}

\begin{figure}
    \centering
        \includegraphics[width=0.3\linewidth]{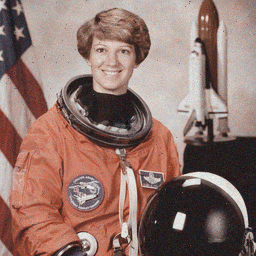}
        \includegraphics[width=0.3\linewidth]{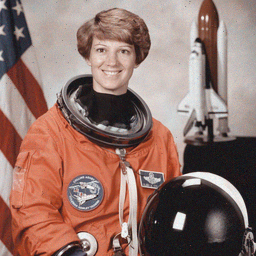}
        \includegraphics[width=0.3\linewidth]{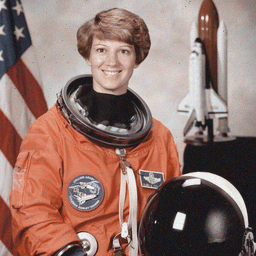}
    \caption{\textbf{Image comparison.} Left to right: random normal init, our MC init, our init.}
    \label{fig:image-comp}
\end{figure}

We compare image reconstruction with Gaussian activation with the three different types of initializations in \cref{fig:image-comp}.

\section{Audio Reconstruction Implementation details}
The audio reconstruction results presented in the main paper differed from the image and SDF reconstruction setups in several key aspects. Specifically, the network architecture used three hidden layers, each containing 256 elements, and the bias terms were initialized identically to the weights. These modifications were consistently applied across all initialization methods examined and proved essential for achieving convergence.

\section{Improvement gap dependence on activation function parameters}
\begin{figure}
    \centering
    \includegraphics[width=0.92\linewidth]{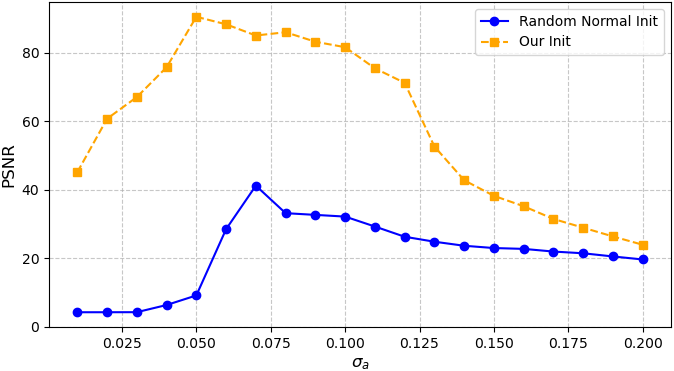}
    \caption{\textbf{Performance gap vs. $\sigma_a$.} As $\sigma_a$ decreases, performance drops for both inits, but a significant gap remains.}
    \label{fig:low_sigma_plot}
\end{figure}

We give results for small $\sigma_a$ in \cref{fig:low_sigma_plot} Left. The results show that the proposed initialization outperforms random init for smaller $\sigma_a$ values while a degradation in performance is observed for both. 

\end{document}